\documentclass[journal,twoside]{IEEEtran}
\usepackage[numbers]{natbib}
\usepackage{amsmath,amssymb}
\usepackage{amsfonts}
\usepackage{algorithm2e}
\usepackage{algpseudocode}
\usepackage{color}
\usepackage{url}
\usepackage{float}
\usepackage{bbm}
\usepackage{graphicx}
\usepackage{verbatim}
\usepackage{multirow}
\usepackage{amsmath}
\usepackage{amssymb}
\providecommand{\keywords}[1]{\textbf{\textit{Index terms---}} #1}

\begin{document}

%
\title{ScoreGAN: A Fraud Review Detector based on Regulated GAN with Data Augmentation}

\title{ScoreGAN: A Fraud Review Detector based on Regulated GAN with Data Augmentation}
\author{Saeedreza Shehnepoor*, Roberto Togneri, Wei Liu, Mohammed Bennamoun
\thanks{S. Shehnepoor (*corresponding author) is with the University of Western Australia, Perth, Australia.
R. Togneri is with the University of Western Australia, Perth, Australia.
M. Buneman is with the University of Western Australia, Perth, Australia.
W. Liu is with the University of Western Australia, Perth, Australia.
emails: \{saeedreza.shehnepoor@research.uwa.edu.au, roberto.togneri@uwa.edu.au, wei.liu@uwa.edu.au, mohammed.bennamoun@uwa.edu.au.\}}
} 
\maketitle




%

%
\begin{abstract}
The promising performance of Deep Neural Networks (DNNs) in text classification, has attracted researchers to use them for fraud review detection. However, the lack of trusted labeled data has limited the performance of the current solutions in detecting fraud reviews. 
The Generative Adversarial Network (GAN) as a semi-supervised method has demonstrated to be effective for data augmentation purposes. The state-of-the-art solutions utilize GANs to overcome the data scarcity problem. However, they fail to incorporate the behavioral clues in fraud generation. Additionally, state-of-the-art approaches overlook the possible bot-generated reviews in the dataset.  Finally, they also suffer from a common limitation in scalability and stability of the GAN, slowing down the training procedure. 
In this work, we propose ScoreGAN 
for fraud review detection that makes use of both review text and review rating scores in the generation and detection process. Scores are incorporated through Information Gain Maximization (IGM) into the loss function for three reasons. One is to generate 
score-correlated reviews based on the scores given to the generator.
Second, the generated reviews are employed to train the discriminator, so the discriminator can correctly label the possible bot-generated reviews through joint representations learned from the concatenation of GLobal Vector for Word representation (GLoVe) extracted from the text and the score.
Finally, it can be used to improve the stability and scalability of the GAN. 
Results show that the proposed framework outperformed the existing state-of-the-art framework, namely FakeGAN, in terms of AP by 7\%, and 5\% on the Yelp and TripAdvisor datasets, respectively.

\keywords fraud reviews detection, deep learning, generative adversarial networks, joint representation, Information Gain Maximization. 
\end{abstract}

\section{Introduction}
\label{sec:intro}
Social media is full of users' opinions on different matters such as news, personal events, advertisements, and businesses. Opinions concerning businesses can greatly influence prospective customers' decisions on purchasing certain products or services. A study in 2015 demonstrated that about 70 percent of people in the US, visit reviews of products, before purchasing\footnote{\url{https://www.mintel.com/press-centre/social-and-lifestyle/seven-in-10-americans-seek-out-opinions-before-making-purchases}}. The openness of popular review platforms (Amazon, eBay, TripAdvisor, Yelp, etc.) provides an opportunity for marketers to promote their own business or defame their competitors, by deploying new techniques such as bots, or hiring humans to write fraud reviews for them. The reviews produced in this way are called ``Fraud Reviews" \cite{Lee2014,Motoyama2011,Wang2012}. Studies~\cite{Zervas2016} show that fraud reviews increased in Yelp by 5\%
to 25\% from 2005-2016. It is worth mentioning that there exists fraud contents in different contexts of social media with the same characteristics \cite{Gong2020}. Fake news consists of articles intentionally written to convey false information for a variety of purposes such as financial or political manipulation \cite{Zhou:2019:FNF:3289600.3291382,Wang17}. Enough knowledge of political science, journalism, psychology, etc. is required to study such contents in social media \cite{Conroy:2015:ADD:2857070.2857152,Shu:2017:FND:3137597.3137600}.

Since the first work on social fraud reviews by Jindal \textit{et al.} in \cite{Jindal2008}, many researchers proposed different approaches to address fraud reviews. The approaches use text based features which extract features from text \cite{Ott2011} (e.g. language models~\cite{Shebuit2015}), and behavioral ones which extract behavioral clues from ``metadata" or reviewers' profiles~\cite{Mukherjee2013}. Text-based and behavioral features when combined have been shown to achieve better performance \cite{Shebuit2015,Shehnepoor2017}. These hand crafted features are fed to classifiers such as the Multi-Layer Perceptron (MLP), Naive Bayes, Support Vector Machines (SVMs) to classify if a review is genuine or not. We call these approaches using hand-crafted feature the \textit{classical} approaches. Recent years have seen Deep Learning (DL) that maps the fraud detection problem to a text classification task, for better feature representation, and to address the overfitting problem \cite{Quoc2014}. 
To deal with data scarcity, a recent attempt~\cite{Aghakhani2018} adopted GAN in a framework called \textit{FakeGAN}. \textit{FakeGAN} consisted of a generator to generate fake reviews and two discriminators. One discriminator is to separate fake from real samples; and the other for discriminating fraud human reviews and fraud generated ones. Despite \textit{FakeGAN}'s simplicity and effectiveness, it suffers from several limitations.
\textbf{First,} although \textit{FakeGAN} generates reviews to address the lack of data, the main limitation is that generated data is not associated with nor exploits any metadata used to extract behavioral features. 
Reviews generated by \textit{FakeGAN}~\cite{Aghakhani2018} provide only review text not correlated with any metadata such as rating scores. Metadata has been shown to be more effective than review text alone in fraud detection~\cite{Shebuit2015,Shehnepoor2017}. 
Generating reviews to correlate with the rating scores provides better representations of reviews learned jointly from both text and metadata.
\textbf{Second,} the previous works also overlooked the possible presence of bot-generated reviews. Bear in mind, reviews in social platforms can be generated automatically by software (a.k.a. bots) trained to generate reviews indistinguishable from human written reviews~\cite{Yuanshun2017}. Deep Neural Networks (DNNs) are employed to produce contents, highly similar to written reviews~\cite{Yuanshun2017}. For such bot-generated reviews, the classical metadata-based methods could result in misleading behavioral features, since bots leave no trace of human activities in the relevant metadata attributes~\cite{Yuanshun2017}.
\textbf{Finally,} \textit{FakeGAN} suffers from instability in the training process. In other words, the training procedure in \textit{FakeGAN} takes a long time to stabilize. Regularizing the objective function is one way to ensure the convergence of a GAN. The semi-supervised approaches also suffer from the lack of scalability on different datasets. 
Experiments on datasets from different domains are required to ensure the scalability of the proposed approach.
In this paper, we propose \textit{ScoreGAN}; a Generative Adversarial Network (GAN) able to generate score-correlated reviews.
To ensure the generation of more meaningful and authentic reviews, we use a concept from Information Gain Maximization theory (IGM)~\cite{Chen:2016:IIR:3157096.3157340} to select generated samples that have the highest information gain against the scores of the reviews. Employing IGM results in generating human like data with score-correlated reviews. IGM was also proposed in~\cite{Chen:2016:IIR:3157096.3157340} in a framework named \textit{InfoGAN}. The basic idea of \textit{InfoGAN} is to generate handwritten numbers using GAN. \textit{InfoGAN} uses information gain maximization to generate image samples with respect to a constraint such as the angle and thickness of a digit's stroke. This idea of incorporating constraints is also applicable to reviews, for example the sentiment (score) of reviews can be considered as a constraint.
To ensure the generation of score-correlated reviews two discriminators are used. One is to discriminate between genuine and fraud reviews ($D_g$). The other discriminator is to ensure the generation of human like fraud reviews ($D_f$). To this end, two softmax layers are used in the second discriminator. One softmax is used to discriminate real fraud and generated fraud reviews through jointly learning the Word Embeddings (WE) of reviews and scores. The second softmax is used to calculate the probability of the score based on the input features. This second softmax theoretically implements the auxiliary function in IGM and is employed in the GAN objective function.  Maximizing  the auxiliary function improves the generation of score-correlated reviews. Note that our purpose of using two discriminators is different from \textit{FakeGAN}. \textit{FakeGAN} uses two discriminators to overcome the ``mode collapse" problem, but here we extend the concept of each discriminator to support IGM.

As a result, generating new score-correlated reviews helps $D_g$ to expand during the learning the joint representations of the text and score from the augmented data.  Generating new data not only provides more diverse data but also allows $D_g$ to train on augmented data. Such augmented data is also produced and propagated throughout social platforms. As explained, employing traditional metadata-based methods could result in misleading behavioral features, since the bots are trained to leave no trace of human activity~\cite{Yuanshun2017}. The metadata constraint in \textit{ScoreGAN} can be selective, because including more metadata attributes can be computationally costly. Furthermore metadata attributes that capture human activity will not be useful when dealing with bot generated reviews. Finally, selecting more than one feature does not necessarily improve the performance of the proposed approach~\cite{Shehnepoor2017}. So in this research we select rating score which has been shown to be the most promising feature~\cite{Shebuit2015,Shehnepoor2017} as the only feature for generating and detecting reviews to avoid misclassification of the possible bot-generated reviews in the testing stage.  
Below we summarize our contributions as follows:
\begin{itemize}
\item We propose ScoreGAN, a novel approach to generate reviews with specific semantics to address the lack of high-quality data in fraud detection. Our results show that customization of the generated reviews based on the score, leads to a significant improvement in fraud review detection by 7\% on the Yelp dataset and 5\% on TripAdvisor as compared with the state-of-the-art systems (See Sec. \ref{subSubSec:performance}, and  \ref{sec:ablative}.`Effect of Score').

\item Fraud reviews in real datasets are either written by fraudsters or generated by bots. Recent approaches in fraud detection rely on text-based and behavioral features which struggle to detect bot generated reviews, which escape detection through manipulation of metadata and written text. For the first time, in this research, we select the generated review candidates from the generator augmenting the data used to train $D_g$. We show that the performance of the fraud detection system can be improved by generating human like reviews and then training the system with the augmented data (See Sec. \ref{subSubSec:performance}. Impact of generated fraud reviews). Then we show that the proposed score-correlated joint representations are the most effective in dealing with bot generated reviews (See Sec. \ref{sec:effects-of-behavioral-features})

\item We perform a comprehensive study on the \textit{stability} and \textit{scalablity} of the \textit{ScoreGAN}. For stability, our experiments demonstrate that adding the regularization term to the objective function improves the convergence of \textit{ScoreGAN} by reducing the number of required epochs (Sec. \ref{sec:ablative}.`Effect of Regularization'). For scalability, we demonstrate that using \textit{ScoreGAN}  to generate labelled data, addresses the data scarcity problem, one of the main challenges in fraud review detection. We show that using a smaller subsets of data, will benefit from generated reviews and converge to the same performance as the full datasets; Both Yelp and TripAdvisor are used to verify the scalability imporvement of \textit{ScoreGAN} (Sec. \ref{sec:robustness}).
\end{itemize}
\section{Related Works}
\label{sec:rel-works}
\subsection{Fraud Detection}
\label{secSub:fraud-detection}
Approaches for fraud detection can be broadly categorized into \textit{Classical} and \textit{Deep Learning (DL)} approaches. 
\subsubsection{Classical Approaches}
\label{secSubSub:detection-classical}
Two types of features are used in fraud detection in this context; behavioral features and text based features \cite{Mukherjee2013}. These features are either used separately or in combination.\\
\textbf{Text-based features:} Text-based approaches extract features directly from the review text~\cite{Myle2012}. Chang \textit{et al.}~\cite{Chang2015} employed pairwise features to detect fraudsters (pairwise features are features extracted by comparing pairs of reviews). Chang \textit{et al.} used similarity among reviews to detect fraud reviews written by group of fraudsters. Previous studies have also shown the significance of n-gram models to improve the accuracy of fraud and fraudster detection~\cite{Song2012}. For example, fraudsters tend to dominate their reviews and as a result use more first person pronouns to increase their impression. In addition to making the reviews bolder, fraudsters use CAPITAL words, to catch the attention of the reviewers \cite{Li2011}.\\
\textbf{Behavioral features:} Behavioral features were initially proposed to address the limitations of the text-based features to capture the behavior of fraudsters. Some of the important behavioral features are reported in~\cite{Jindal2008,Shebuit2015,Shehnepoor2017}. As an example, writing reviews on every hotel in a town is unusual, since a traveler will likely use just one hotel in a specific town~\cite{Chavoshi2015}. Fraudsters also tend to write as many reviews as they can, since they are paid based on the number of reviews they write. So as the number of reviews for a certain reviewer increases, the probability for him/her to be a fraudster increases~\cite{Arjun2012}. In addition, normal users have a low deviation of their opinions 
while fraudsters tend to promote the services for the companies they are working for, and defame the services of the competitors.
So a user's score can also be considered a behavioral feature of a fraudster~\cite{Ee-Peng2010}. 
NetSpam~\cite{Shehnepoor2017} employed both text and behavioral features in four categories providing 8 different types of features. These features were fed to a Heterogeneous Information Network (HIN) to output a ranked list of reviews based on the probability to be fraud. NetSpam was evaluated on the Yelp and the Amazon dataset as real world datasets to demonstrate the scalability of the proposed approach.
\subsubsection{Deep Learning}
\label{secSubSub:detection-DL}
In recent years, Deep Learning (DL) has attracted attention for different purposes for two main reasons. \textbf{First}, hidden layers in DL neural networks are able to extract complex hidden information in sentences. \textbf{Second}, a global representation of a sentence is achievable using such networks. Hand crafted features fail to do both~\cite{Quoc2014,Duyu2015}. 
Ren \textit{et al.}~\cite{Yafeng2016} used DL to detect fraud reviews. Ren \textit{et al.} employed a Convolutional Neural Network (CNN) to extract a representation of a sentence from Word Embedding (WE). Then the sentence representation is fed to a Gated Recurrent Neural Network (GRNN) and generates a document representation of the review. The features are then fed to a softmax layer to determine if the review is fraud or genuine. This approach demonstrated a 3\% improvement in fraud classification, on TripAdvisor. The term frequency, word2vec and LDA (Latent Dirichlet Allocation) were combined by Jia \textit{et al.}~\cite{Shaohua2018} to spot fraud reviews on Yelp. 
Jia \textit{et al.} extracted 5 topics from fake and non-fake reviews and described each topic using 8 words. The features were then fed to a SVM classifier, Logistic Regression and Multi-Layer Perceptron (MLP) and the results show that the MLP achieved the best performance (81.3\% Accuracy).
Li \textit{et al.}~\cite{li2020detection} introduced a new concept called review groups (an extension to the reviewer groups which refers to a group of reviewers cooperating to collectively promote/demote products). Li \textit{et al.} claim that reviews also can be grouped based on the assumption that the reviews in the same group share similar credibility and thus the same possibility to be fraud/genuine. In the first step, Word Embeddings (WE) were learned through Continous Bag of Words (CBoW) combined with four metrics: review posting time, review rating, store category and review content emotion to calculate the distance between every two reviews.  A threshold was applied to group the reviews.  Next, the WE of the reviewers in a group was combined with different group behavioral features  (e.g., inner group time density, inner group store similarity,  and inner group rate diversity) to form the final representation of the review group. The final representation was fed to different classifiers (e.g., SVM, Naive Bayes, and Random Forest) to label the review groups.
Shaalan \textit{et al.}~\cite{shaalandetecting} proposed a two-step approach concerning the key features of fraudsters such as adding redundant information in written review text or writing reviews in bursts.  In the first step,  a  Deep  Boltzmann Machine (DBM) was used as the aspect-level sentiment model. In the second step, a Long Short Term Memory (LSTM) was applied to the sentiment aspect-level representation extracted for each review. The output of the LSTM was the label of a review to be fraud/genuine.
\subsection{Generative Adversarial Networks}
\label{secSub:GAN}
Generative Adversarial Networks (GANs) \cite{Goodfellow2014a} are among the latest approaches that have been used in Artificial Intelligence (AI) for different applications.
Liang~\textit{et al.}~\cite{Liang2017} produce a description of an image using the generator \textit{G} and the discriminator \textit{D} distinguishing between the generated description and the real description. Liang \textit{et al.} claimed that the insufficient labelled descriptions may result in overfitting, and the GAN can generate new descriptions used to both augment and also provide real descriptions of images. 

Aghakhani~\textit{et al.}~\cite{Aghakhani2018} proposed \textit{FakeGAN} to investigate the problem of fraud review detection using a GAN, by generating fraud reviews with the generator and then the GAN is fed with review text. \textit{FakeGAN} consists of a generator to generate fake samples as auto-generated bot reviews and two discriminators. One to discriminate between fake and real samples and the other to discriminate fraud human reviews and fraud generated ones. Unlike the original GAN, \textit{FakeGAN} utilizes the dual discriminators to overcome the well-known ``mode collapse" problem. Mode collapse refers to a situation that the generator switches between different modes during the training phase, because of the complexity of the input data distribution. \textit{FakeGAN} was evaluated on the TripAdvisor dataset containing 800 reviews, 400 real and 400 deceptive obtained from~\cite{Ott2011}.

\section{Proposed Methodology}
\label{sec:method}
\subsection{Problem Definition}
\label{secSub:problem-definition}
Given a set of real reviews $X$, consisting of genuine reviews and corresponding scores $\langle X_g,s\rangle$ and fraud human reviews and corresponding scores $\langle X_{fh},s\rangle$, our purpose is to design a system that generates a set of score-correlated fraud bot reviews $\langle X_{fb},s\rangle$. We denote the fraud reviews as $X_f = \{X_{fh}, X_{fb}\}$. First, we train a discriminator $D_f$ to differentiate $X_{fh}$ from $X_{fb}$ and calculate the probability of a score based on $\langle X_f,s\rangle$. This will ensure that we generate more human like fraud reviews $X_{fb}$, which in turn allows us to train the discriminator $D_g$ to differentiate genuine reviews ($X_{g}$) from fraud ones ($X_{f}$). 

\begin{figure}
  \centering
  \includegraphics[width=\linewidth]{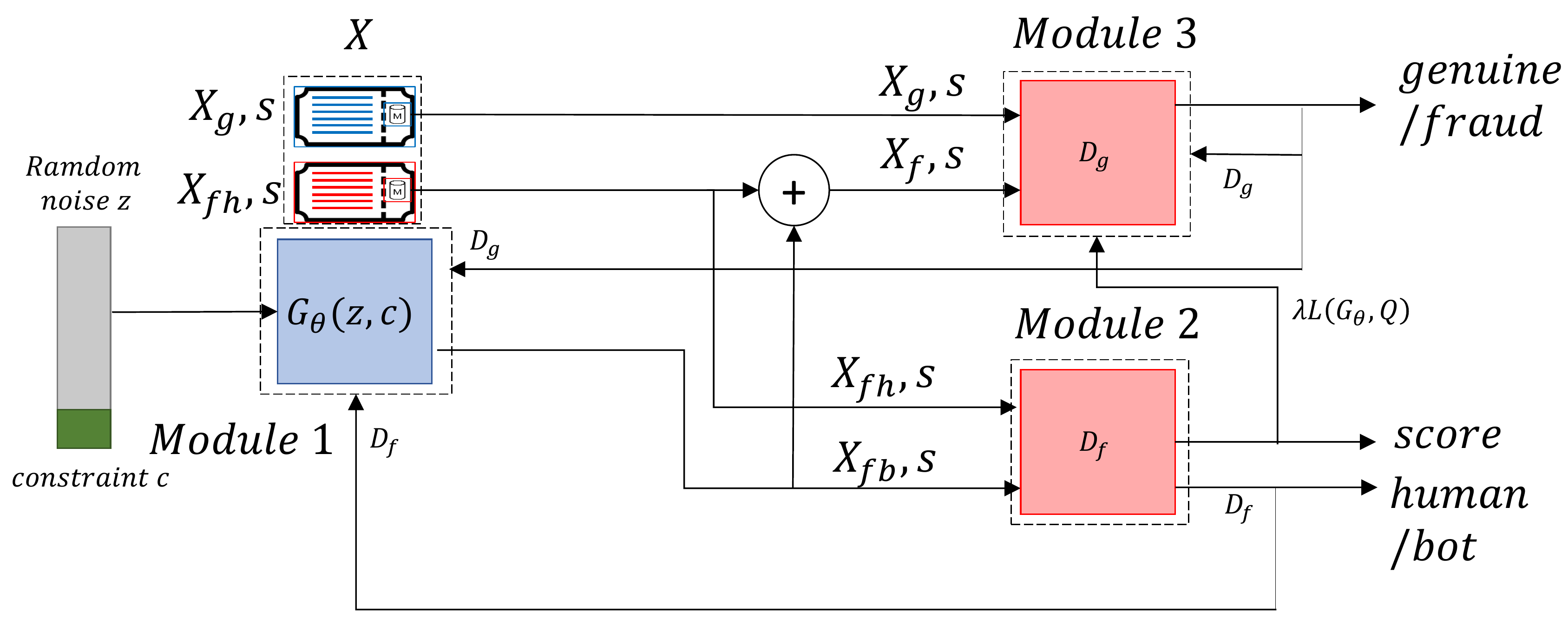}
  \caption{Framework of ScoreGAN. Note $\oplus$ represents concatenation.}
    \label{fig:framework}
\end{figure}

Fig. \ref{fig:framework} depicts the overall system architecture of ScoreGAN, our proposed regularized GAN based fraud review detection system with regularized GAN. In the following, we provide the related explanations to components of ScoreGAN in details. 

\subsection{Information Gain Maximization GAN Regularization}
\label{secSub:gan-constraint}
To generate reviews conditioned on a specific constraint $c$, we need to increase the mutual information between $c$ and the generator model $G_\theta(z,c)$, where $z$ is some random noise. We need to add a regularization term to the objective function to maximize the mutual information gain between $c$ and $G_\theta(z,c)$; namely $I(c,G_\theta(z,c))$. With entropy ($H$) definition, $I(c,G_\theta(z,c))$ is defined as follows:
\begin{equation}
\begin{split}
I(c,G_\theta(z,c))&=H(c) - H(c|G_\theta(z,c)) \\
& = -\mathbf{E}_{x\sim G_\theta(z,c),c\sim P(c|x)}[-\log (P(c|x))]\\
& + H(c) 
\label{eq:Info-gain-ext}
\end{split}
\end{equation}
From Eq. \eqref{eq:Info-gain-ext} it is difficult to maximize the information gain directly, since sampling the posterior $P(c|x)$ is required. Therefore we use variational mutual information maximization \cite{Barber:2003:IAV:2981345.2981371} to find a lower bound over $I(c,G_\theta(z,c))$ to make the lower bound as tight as possible. To do that, we first need to define an auxiliary distribution, $Q(c|x)$, an approximate for $P(c|x)$. So we extend Eq. \eqref{eq:Info-gain-ext}:
\begin{equation}
\begin{split}
& = -\mathbf{E}_{x\sim G_\theta(z,c),c'\sim P(c|x)}[-\log (P(c'|x))] + H(c) \\
& = \mathbf{E}_{x\sim G_\theta(z,c)}[\mathbf{E}_{c'\sim P(c|x)}[\log (P(c'|x))]] + H(c)\\
& = \mathbf{E}_{x\sim G_\theta(z,c)}[D_{KL}(P(.|x)||Q(.|x))] \\
& + \mathbf{E}_{c'\sim P(c|x)}[\log (Q(c'|x))] + H(c) \\
& \geq \mathbf{E}_{x\sim G_\theta(z,c)}[\mathbf{E}_{c'\sim P(c|x)}[\log (Q(c'|x))]] + H(c) \label{eq:Info-gain-proof}
\end{split}
\end{equation}
In Eq. \ref{eq:Info-gain-proof} we are able to maximize the entropy $H(c)$, since the distribution of $c$ is fixed. 
Next, we state a simple lemma (proof in~\cite{Chen:2016:IIR:3157096.3157340}) that removes the need to sample from the posterior.\\
\textbf{Lemma III.1} \textit{For random variables X, Y and function f(x, y) under suitable regularity conditions:\\
$\mathbf{E}_{x\sim X,y\sim Y|x}[f(x,y)] = \mathbf{E}_{x\sim X,y\sim Y|x,x'\sim X|y}[f(x',y)]$}\\
By using the Lemma, we can define a variational lower bound for $I(c,G_\theta(z,c))$.
Using the lemma, we have:
\begin{equation}
\begin{split}
L(G_\theta,Q) & = -\mathbf{E}_{x\sim G_\theta(z,c),c\sim P(c)}[-\log (Q(c|x))] + H(c) \\
& = \mathbf{E}_{x\sim G_\theta(z,c)}[\mathbf{E}_{c'\sim P(c|x)}[\log (Q(c'|x))]] + H(c) \\
&\leq I(c,G_\theta (z,c))
\end{split}    
\end{equation}
To incorporate $L(G_\theta,Q)$ in review generation, we add a fully connected layer to the output parameters of $D_f$ to calculate $Q_s$, which models the probability of a review to have the score $s$. Hence, we can regulate the objective function of the GAN to solve the minmax game for $D_g$ as follows:
\begin{equation}
\begin{split}
    \max( \mathbf{E}_{x\sim X_g}[\log D_g(x)] 
    & + \mathbf{E}_{x\sim X_{fh} }[1-\log D_g(x)] \\
    &  + \lambda L(G_\theta,Q)) 
    \label{eq:obj-fun-new}
\end{split}    
\end{equation}
In addition to training GAN, the objective function of Eq. \ref{eq:obj-fun-new} aims to preserve the contribution of $c$ during the generation process. 
    
\subsection{Models}
\subsubsection{Generator (Module 1)}
\label{secSubSub:generator}
\begin{figure}[H]
  \centering
  \includegraphics[width=\linewidth]{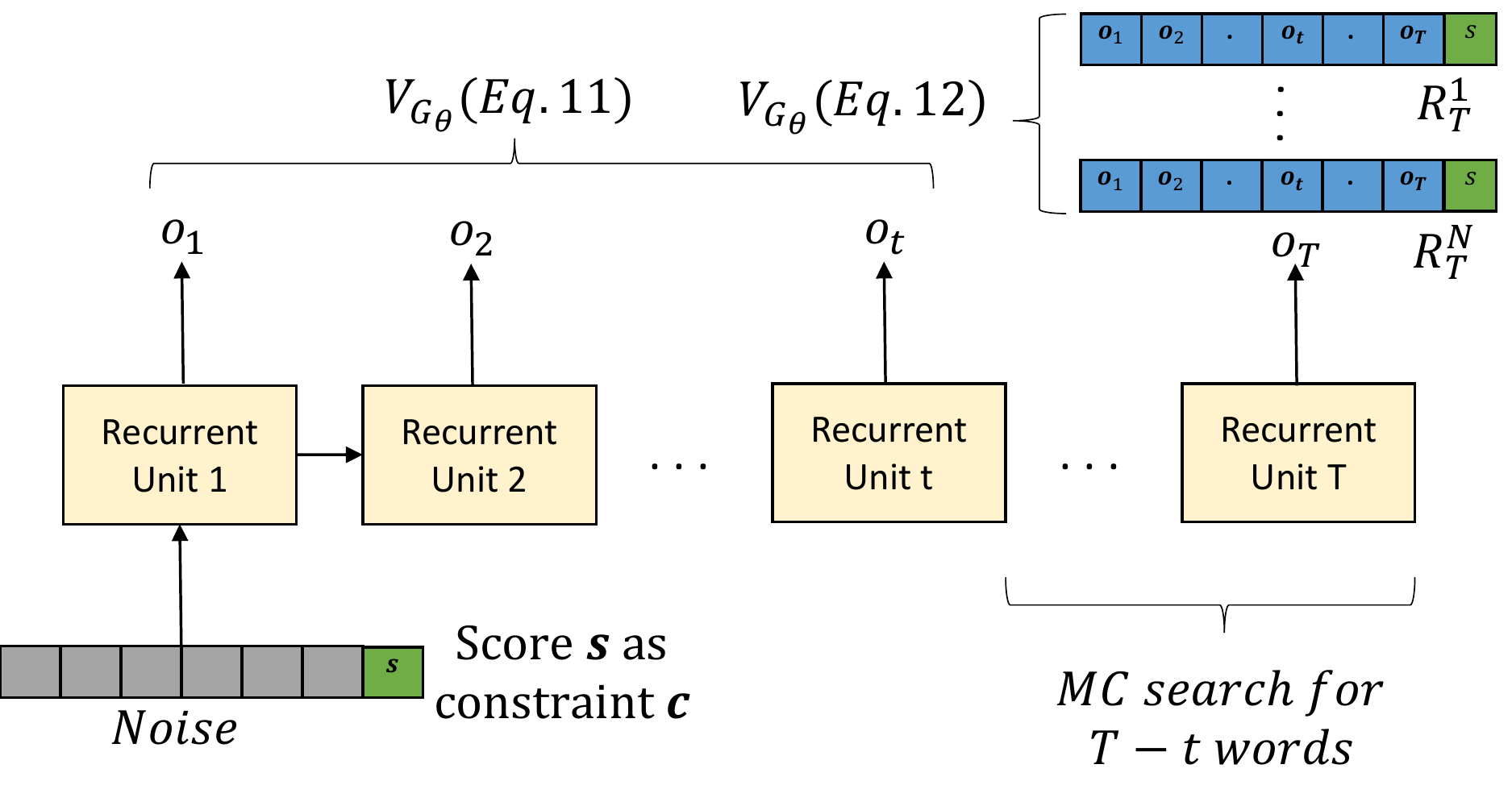}
  \caption{Generator architecture.}
    \label{fig:generator}
\end{figure}

The overall architecture of the generator is depicted in Fig. \ref{fig:generator}. The generator takes the random noise $z$ and score $s$ as a constraint, and generates fake bot reviews $X_{fb}$. Note that in the original  GAN, there is no constraint on the generated samples. Here, the generator generates samples that satisfy  $P_{G_\theta}(x|c)$ not just $P_{G_\theta}(x)$. $P_{G_\theta}(x|c)$ denotes the probability that the generator $G$ generates a sample $x$ given constraint $c$. $P_{G_\theta}(x)$ is the probability that the generator $G$ generates a sample $x$ without any constraints. To generate the words, we used a  Long  Short  Term  Memory  (LSTM),  where each recurrent unit has embedding size and a hidden dimension of 32. 
The  LSTM is trained with a batch size of 50.
We fed the noise concatenated with the score $s$ to the LSTM as an input and produced the hidden states:
\begin{equation}
\label{eq:hidden-state}
    h_t = g(h_{t-1},o_{t-1})
\end{equation}
In Eq. \ref{eq:hidden-state}, $o_t$ represents the sequence of the generated word embedding which is mapped into a sequence of hidden states, $h_t$, using gate function $g$. The output was generated using a softmax function.
\begin{equation}
    \label{eq:softmax}
    o = \frac{e^{W_th_t + b}}{\sum_{t' = 1}^{T}e^{W_{t'}h_t + b}}
\end{equation}
where $o$ is the output of the softmax function, $W$ is the weight matrix and $b$ is the bias vector. 

\subsubsection{Discriminators (Module 2, 3)}
\begin{figure*}
  \centering
  \includegraphics[width=\linewidth]{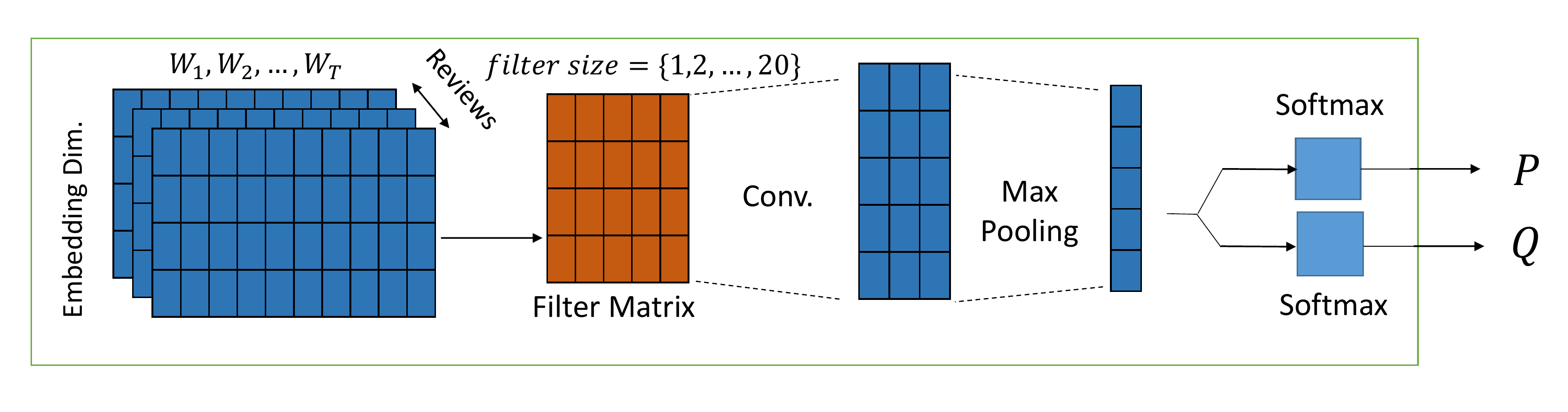}
  \caption{Discriminator architecture. Note that $D_g$ only employs $P$, while $D_f$ also add $Q$ to implement the auxiliary function explained in Sec. \ref{secSub:gan-constraint}.}
    \label{fig:discriminator}
\end{figure*}
\label{secSubSub:discriminator}
Fig. \ref{fig:discriminator} represents the details of the discriminators. Two discriminators are used in ScoreGAN, one ($D_f$) is to help the generator to generate score-correlated reviews, and the other one ($D_g$) is to discriminate between fraud and genuine reviews. For training both discriminators we used 2D convolution filters with different filter sizes ranging from 1 to 20  and 100, 160,  and  200  filter numbers. The filters have a shape of $[filter\_size\times embedding\_size\times filter\_number]$. After each filter, a max-pooling is applied and the obtained feature representations are concatenated and fed to a softmax layer for final classification. Note that $D_g$ only uses one softmax to label genuine/fraud reviews. On the other hand, $D_f$ uses two softmax: one to classify the reviews as real fraud/generated fraud reviews ($P$); the other softmax ($Q$) calculates the probability of each score (ranging from 1 to 5) based on input features,  as an implementation of the auxiliary function introduced in Sec. \ref{secSub:gan-constraint}. 
We represent each review as $\{e_{w_1}, ..., e_{w_t}, ..., e_{w_T}\}$, where $e_{w_t}$ is the embedding of word $t$. A CNN was used as the classifier for fraud review detection \cite{khan2018}. We first define a concatenated word embedding as:
\begin{equation}
    \label{eq:concat}
    e_{1:T} = e_{w_1} \oplus ... \oplus e_{w_t} \oplus ... \oplus e_{w_T}
\end{equation}
where $\oplus$ symbol represents concatenation.
We apply a convolution layer to a window size of $u$ using an ReLU function:
\begin{equation}
\begin{split}
m = ReLU(k\cdot e_{1:1+u-1} + b) 
& \oplus ReLU(k\cdot e_{2:2+u-1} + b)\oplus ...\\  
& \oplus ReLU(k\cdot e_{T-u+1:T} + b)\oplus s
    \label{eq:ReLU}
\end{split}
\end{equation}
where $k$ is the kernel function, $u$ is the window size, and $b$ represents a bias value. The representation is then concatenated with the review score $s$. The output for this step is $m$ as a vector representation for the review. A max-pooling is applied to $m$ to get the combination of the different kernel outputs. The max-pooling output is $f$.

Finally, a softmax function was applied to $f$ to calculate the class probability $P$, of the fraud and genuine reviews:
\begin{equation}
    \label{eq:d-softmax}
    P_i = \frac{e^{W^{P}_if+b^P}}{\sum_{l = 1}^{L}e^{W^{P}_lf + b^P}}
\end{equation}
where $P_i$ is the probability of the review to be labelled as $i$ (fraud/genuine) and $W^P$ is the weighting matrix for the softmax $P$. The $b^P$ represents the bias value and the $L$ is the number of the review labels (fraud/genuine).  
As previously explained (and depicted in Fig. \ref{fig:discriminator}), $D_f$ also uses another softmax function to calculate $Q$ to enforce an update on the constraint in the cost function, using the following function:
\begin{equation}
    \label{eq:c-softmax}
    Q_s = \frac{e^{W^{Q}_sf+b^Q}}{\sum_{c = 1}^{C}e^{W^{Q}_cf + b^Q}}
\end{equation}
where $Q_s$ is the probability of the review to have score $s$ used in Eq. \ref{eq:obj-fun-new} for the regularization term. The $W^Q$ is a weighting matrix trained based on different scores, and the $b^Q$ is the bias vector. 
The $C$ represents the number of different scores ranging from 1 to 5.
\subsection{Training}
Training the ScoreGAN consists of two steps; pre-training and training. Pre-training is used to generate first reviews for the subsequent training of the $D_f$ with $X_{fh}$ and $X_{fb}$. For training, we use a Monte-Carlo search, to solve the problem of discrete token generation for the generator. Monte-Carlo is a search algorithm to identify the most promising moves in a game, heuristically. For each state, the algorithm plays the game to the end for a fixed number of times, based on a specific policy. The best move is selected based on a reward for a complete sequence of moves, where each move is a selected word in this study.
\subsubsection{Pre-training}
In the pre-training section, we need to generate some fake bot reviews  $X_{fb}$ as an input for $D_f$. So we first train the $G_\theta(z,c)$ on fake human reviews $X_{fh}$ using Maximum Likelihood Estimation. The discriminators are also pre-trained using cross-entropy with generated reviews and fake reviews as input for $D_f$, while fraud reviews and genuine reviews as input for $D_g$.
\subsubsection{Adversarial Training}
We adopt the idea of the roll-out policy model for reinforcement learning to generate the sequence of words for $X_{fb}$. So in adversarial training, we aim to generate samples with higher rewards (more realistic) from $D_g$ and $D_f$, and also consistent with the scores (specific sentiment) from $D_f$. This forces the reviews generated by the generator to take actions that lead to a better reward from the discriminator and subsequently higher rewards in the policy gradient of the Monte-Carlo search. So the $D_f + D_g$ indicates the reward or quality of the generated reviews from $G_\theta(z,c)$. The action-value for a taken action $a$ considering state $s$ by the generator is calculated by:
\begin{equation}
    \label{eq:action-value}
    V_{G_\theta}(a = o_t, s = o_{1:t-1}) = D_f(o_{1:t}) + D_g(o_{1:t})
\end{equation}
The problem here is that in $G_\theta(z,c)$, every generated word for the final review generation needs a reward and discriminators can only calculate the reward for fully generated genuine or fraud generated sentences and not the incomplete ones. 
Therefore, for reviews with length less than a complete sentence ($T$ as the length for a complete review and $t$ for the arbitrary length) we need to perform a Monte-Carlo search on words to predict the remaining ones. For a good prediction, an $N$-time Monte-Carlo search is employed. The reviews generated by $N$-time Monte-carlo is defined as $\{R_{1:T}^1, ..., R_{1:T}^N\}$ which are sampled using roll-out policy $G'_\beta$ based on their current state. It is worth mentioning that $G'_\beta$ is the same generative model we used for generation, hence $G'_\beta = G_\theta$. For the complete review we also consider $D_f+D_g$ is the reward.  For a incomplete sentence, though, it can be calculated using the following equation:
\begin{equation}
    \label{eq:reward}
    V_{G_\theta} = \frac{1}{N}\sum_{j = 1}^{N} (D_f(R_T^j) + D_g(R_T^j))
\end{equation}
where $V_{G_\theta}$ is the reward value of the incomplete generated review $R_T^j$ which is completed by an $N$-times Monte-Carlo search and is rewarded the same number of iterations by two discriminators. \\
Using this approach we can convert the discrete words into continuous form. The updates can be then propagated backwards from discriminators through the generator. Finally, we complete the adversarial training, which is a function to maximize the final reward. We use the following objective function for this purpose:
\begin{equation}
    J(\theta) = \sum_{R_i\in R_{G_\theta(z,c)}} G_\theta(R_i)*V_{R_i}
\end{equation}
So for updating $\theta$ we just need:
\begin{equation}
\label{eq:gen-update}
    \theta \Longleftarrow \theta + \gamma\bigtriangledown_{\theta}J(\theta)
\end{equation}
where $\gamma$ is the learning rate and is set to 1. In addition we use Eq. \ref{eq:obj-fun-new} for training the discriminators. So both the generator and the discriminators are updated mutually to finally converge to an optimum point. Algorithm \ref{alg:GAN-alg} describes how ScoreGAN works.  
 
\begin{algorithm}
 \caption{ScoreGAN Algorithm}
\label{alg:GAN-alg}
\KwResult{reviews probability ranked by $D_g$ as fraud, customized reviews generated by $G_\theta(z,s)$}
 \% Pre-training\;
 generating word embedding of words in $X_{fh}$ and $X_g$\;
 pre-train $G_{\theta}(z,s)$ with word embedding of $X_{fh}$\;
 
 \% Training\;
 \While{$\sim$ convergence}{
  \% Generator training\;
  \For{$it \gets 1$ to $IT$} 
    {\% generating customized reviews based on score\;
    generate $X_{fb}$ from $z$ and score $s$\;
        \For{$t \gets 1$ to $T$}
        {
            \% Reward for each word based on $D_f$ and $D_g$\
            compute $V_R$ by Eq. \ref{eq:reward}\;
        }
        update $\theta$ by Eq. \ref{eq:gen-update}; 
}
  \For{$it' \gets 1$ to $IT'$}
    {
      \% Samples from generator as positive input for $D_f$ and negative one for $D_g$\;
        generate $R_{G}$ \;
        train $D_f$ with $X_{fh}$ as (+) and $X_{fb}$ as (-)\;
        train $D_g$ with $X_g$ as (+) and $X_f$ as (-)
    }

 }
 \% Testing\;
 generate $X_{fb}$ by $G_\theta(z,s)$\;
 compute probability of $X_{fh}$, $X_g$ to be fraud\;
\end{algorithm}

\section{Results and Evaluation}
\subsection{Datasets}
\label{sec:datasets}
As discussed in Sec. \ref{sec:intro}, datasets for fraud review detection labeled by humans are referred to as ``near ground-truth". Most of the existing datasets only provide review text, rather than both text and metadata. We need a labeled dataset containing both text and metadata to meet the requirement of the proposed approach. In this study we use the Yelp and the TripAdvisor datasets. Yelp is a social media platform which provides the opportunity for people to write reviews of their experience of different restaurants and hotels in NYC. The dataset is labeled by the Yelp filtering system, which is more trusted than other datasets labeled by human \cite{Mukherjee2013}. The dataset contains review ID, item ID, user ID, score (rating from 1 to 5) given by different reviewers on different items, date of written reviews, and text itself.    
TripAdvisor provides the opportunity for people to write reviews about different entertainment places and rate them. Unlike the Yelp dataset people are not able to rate the businesses. They can only like or dislike the business based on their negative or positive tendency. The dataset contains the review texts together with the people's tendency in the form of like or dislike, labeled by human judges. Since most of the reviews are less than 400 words, we selected the reviews with less than 400 words. Then we pad words with ``END", so they can have a length of 400. The dataset does not provide any information about the users who wrote the reviews. Using two datasets representing different businesses, we attempt to show both the scalability of our proposed ScoreGAN approach and also its ability in coping with the impact of missing data. Details of the two datasets are listed in Table \ref{tab:datasets}. 

\begin{table*} 
\centering
  \caption{Details of datasets.}
  \label{tab:datasets}
  \begin{tabular}{ccccc}
    \hline
    Datasets & Reviews (spam\%) & Users & Resto. \& hotels & Rating\\ \hline
    Yelp-main & 6,000 (19.66\%) & 47 & 5,046 & 1-5 \\
    TripAdviser & 1600 (50\%) & - & 20 & -1 (dislike), 1 (like) \\ 
    \hline
    \end{tabular}
\end{table*}
\subsection{Experimental Setup}
\label{sec:exp-set}
We used GLoVe~\cite{pennington-etal-2014-glove} as the baseline system for the word embedding with dimension 50, and a batch size of 64 for the inputs of discriminators. For CNN, we used different filter sizes for the hyper-tuning of the results following the same practice of different successful studies of text classification \cite{Lai:2015:RCN:2886521.2886636,Zhang:2015:CCN:2969239.2969312}. The input filters are representative of the n-gram language model, which in our case $n$ is chosen from $\{1,2,...,20\}$. We used the weighting matrix to map the input features (obtained from concatenation using Eq. \ref{eq:concat}) to a one dimension representation with different filter size, chosen from $\{100,160,200\}$. The learning rate for the discriminators is $10^{-4}$ and it is $10^{-20}$ for the generator. The training iterations are set to 100 for the generator and discriminators. For adversarial training, we used 120 iterations. In algorithm \ref{alg:GAN-alg} for each outer loop, the generator was trained 5 iterations in the inner loop ($IT = 5$). The training epochs for the discriminators trained is set to 3 ($IT' = 3$).  
\begin{table*}
\centering
  \caption{ScoreGAN performance, using 70\% of the Yelp-main as a training set and 30\% as a test set.}
\small
  \label{tab:comparison}
  \begin{tabular}{|c|c|ccc|}
    \hline
    Dataset& Framework & AP & AUC & Accuracy \\ \hline
    \multirow{5}{*}{Yelp-main} & NetSpam & 0.5832 $\pm$ 0.0028  & 0.7623 $\pm$ 0.0192  & 0.7232 $\pm$ 0.0293 \\
    &  Li \textit{et al.}~\cite{li2020detection} & 0.6003 $\pm$ 0.0018 & 0.7823 $\pm$ 0.0235 & 0.7795 $\pm$ 0.0291\\
    & Shaalan \textit{et al.}~\cite{shaalandetecting} & 0.6201 $\pm$ 0.0029 & 0.8238 $\pm$ 0.0183 & 0.8085 $\pm$ 0.0148\\
    & FakeGAN & 0.5959 $\pm$ 0.03684  & 0.8686 $\pm$ 0.01334 & 0.8280 $\pm$ 0.0045 \\
    & ScoreGAN & \textbf{0.6516 $\pm$  0.0275} & \textbf{0.8878 $\pm$ 0.02012} & \textbf{0.8476 $\pm$ 0.0048} \\ \hline
    \multirow{5}{*}{TripAdvisor}& Netspam & {0.6194 $\pm$ 0.0093} & {0.7782 $\pm$ 0.0174} & {0.7428 $\pm$ 0.0029} \\
    &  Li \textit{et al.}~\cite{li2020detection} & 0.5966 $\pm$ 0.0093 & 0.7506 $\pm$ 0.0173 & 0.7102 $\pm$ 0.0261\\
    & Shaalan \textit{et al.}~\cite{shaalandetecting} & 0.6494 $\pm$ 0.0041 & 0.8053 $\pm$ 0.0113 & 0.7502 $\pm$ 0.0338\\
    & FakeGAN& {0.6858 $\pm$ 0.0403} & {0.8510 $\pm$ 0.0336} & {0.7619 $\pm$ 0.0461}\\
    & ScoreGAN & \textbf{0.7160 $\pm$ 0.0058} & \textbf{0.8767 $\pm$ 0.0197} & \textbf{0.7726 $\pm$ 0.0084}\\ \hline
\end{tabular}
\end{table*}
\subsection{Main Results}
In this section we evaluate the performance of our proposed system based on three well-known metrics: Average Precision (AP), Area Under Curve (AUC), and Accuracy. To this end, we performed an ablative study on the effect of score, behavioral features impression, and regularization. Then we examine ScoreGAN robustness against data scarcity. Note that all the results are based on the performance of $D_g$ as the main discriminator. 
\subsubsection{Performance}
\label{subSubSec:performance}
\begin{figure*}
  \centering
  \includegraphics[width=\linewidth]{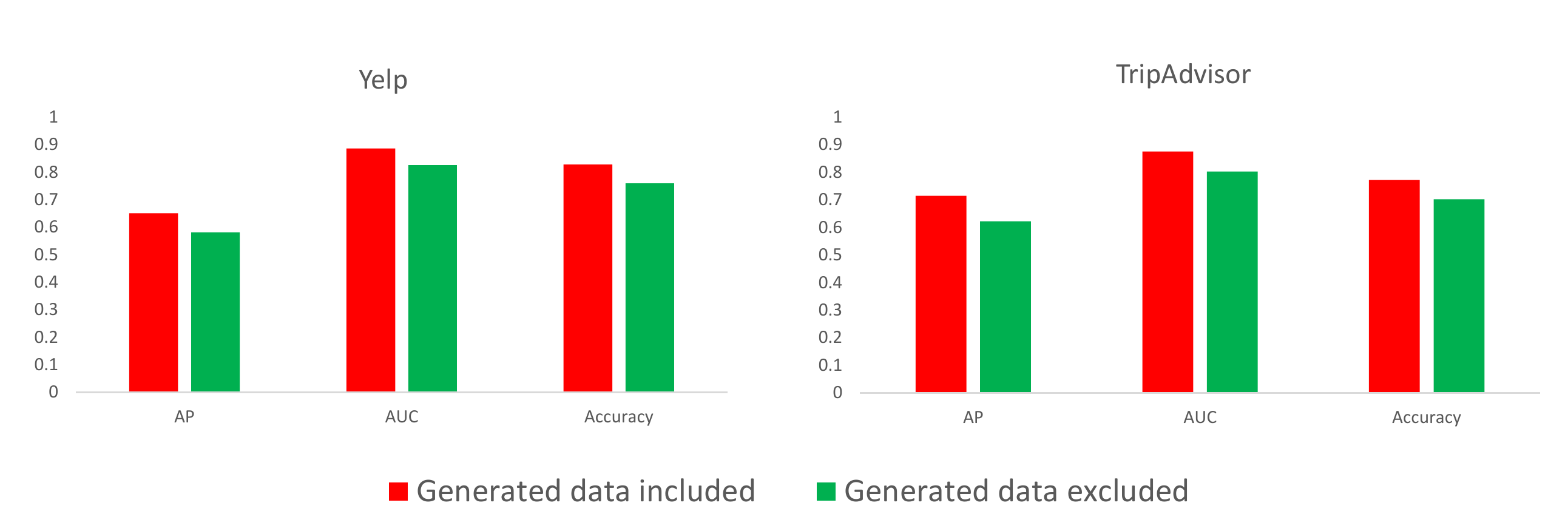}
  \caption{Effect of the generated reviews on $D_g$ performance with 70\% of data as training set.}
\label{fig:bot-impact}
\end{figure*}

We used both datasets from Table \ref{tab:datasets}. 
As Table \ref{tab:comparison} shows, the proposed \textit{ScoreGAN} outperforms other systems for both datasets according to the three metrics. $AP$ is highly dependent on the fraud percentage in the dataset, while $AUC$ is independent of the fraud percentage. The improvement is mostly because \textit{ScoreGAN} combines the key strengths of both \textit{FakeGAN} (synthetic data generation) and \textit{NetSpam} (combination of multiple features). 
The framework proposed by Li \textit{et al.}~\cite{li2020detection} solely relies on review group features (text-based and behavioral), making it unable to detect contents generated by adversarial networks, resulting in lower performance. The approach by Shaalan \textit{et al.}~\cite{shaalandetecting} on the other hand relies only on sentiment analysis of reviews and overlooks the behavioral features.\\
\textit{\textbf{Impact of Generated Fraud Reviews:}}
As mentioned in Sec. \ref{sec:intro}, generated reviews play an important role in improving the performance. Fig. \ref{fig:bot-impact} displays the performance of \textit{ScoreGAN} when it is only trained with human fraud reviews (green) vs. when it is trained with the combination of generated and fraud reviews (red).  
The performance of the \textit{ScoreGAN} is improved over all of the metrics when real data is combined with generated data. Generated reviews augment the training data which leads to better performance. In addition, generated reviews can imitate the bot written reviews in the datasets, helping the discriminators to learn the more diverse data, enabling discriminators to detect the bot reviews in real datasets with a better performance.

\subsubsection{Behavioral Feature Performance on ScoreGAN}
\label{sec:effects-of-behavioral-features}
\begin{figure}
  \centering
  \includegraphics[width=\linewidth]{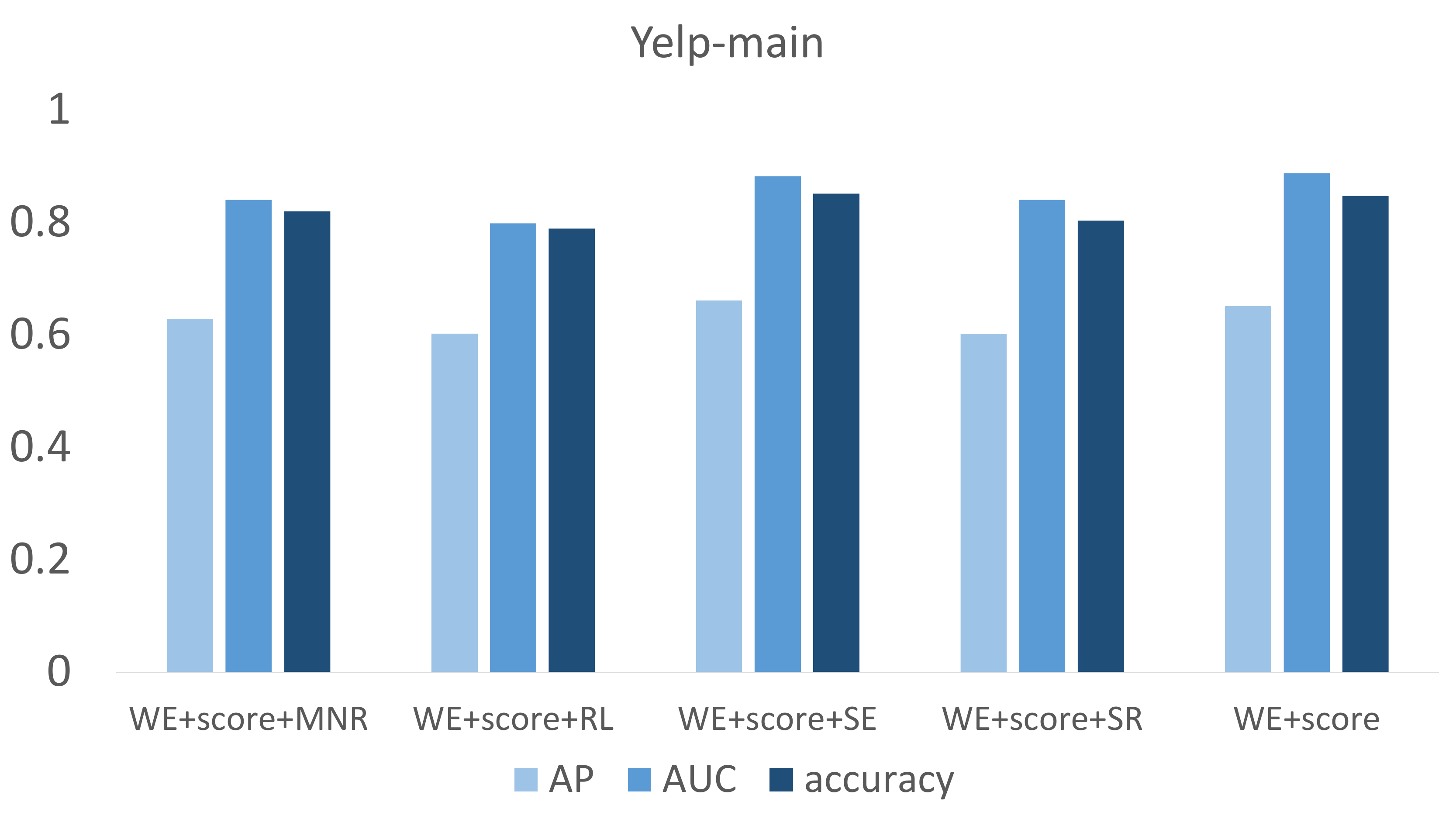}
  \caption{The performance of ScoreGAN on the Yelp dataset with different combinations of WE and behavioral features.}
\label{fig:beh-yelp}
\end{figure}

\begin{figure}
  \centering
  \includegraphics[width=\linewidth]{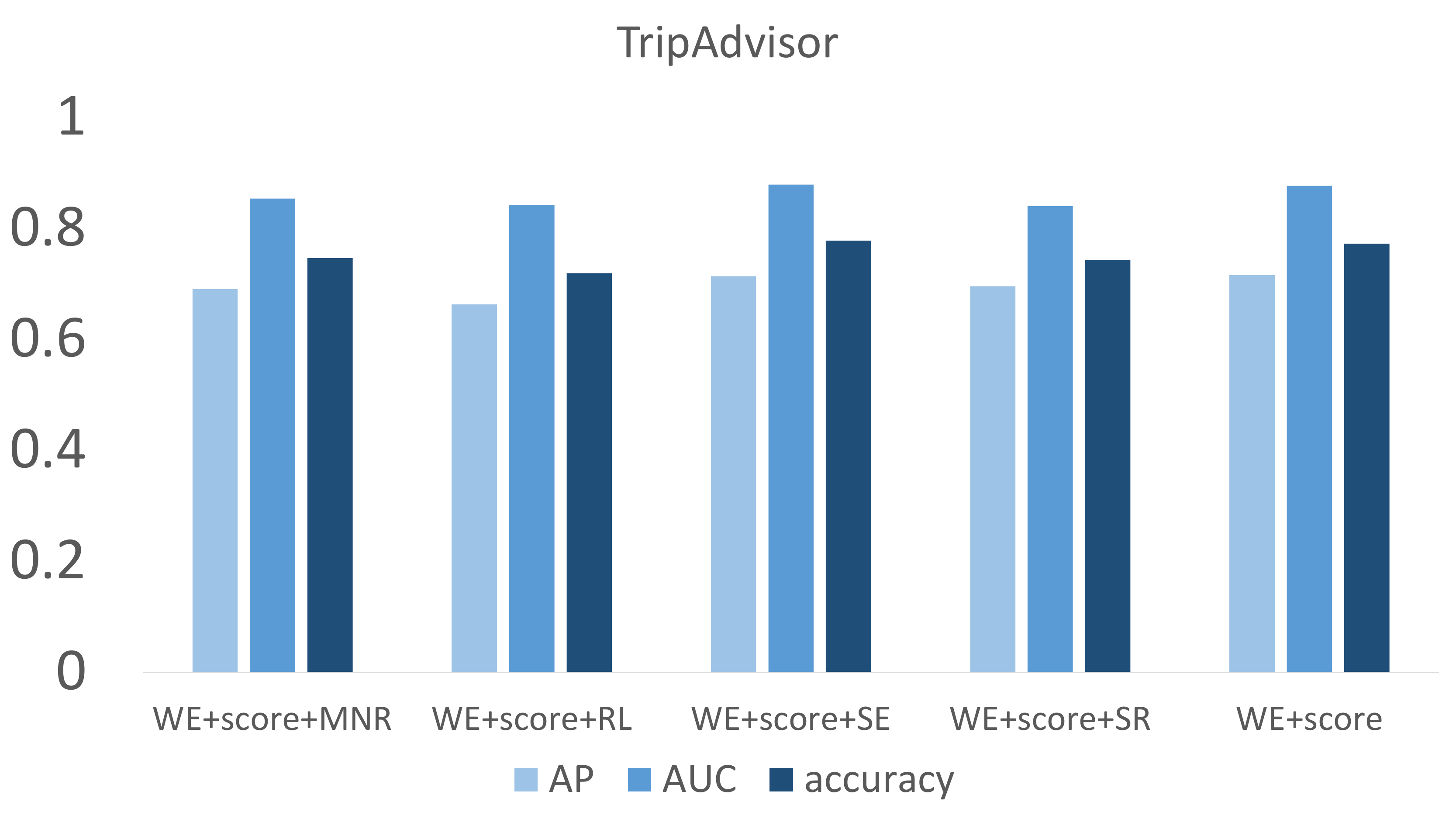}
  \caption{The performance of ScoreGAN on the TripAdvisor dataset with different combinations of WE and behavioral features.}
\label{fig:beh-trip}
\end{figure}

Behavioral features play an important role in identifying fraud reviews in social media. To investigate the effectiveness of the behavioral features we devised an experiment with different behavioral features concatenated with WE as the vector representation for each review.  Four  different  behavioral  features are  used  in this study to  evaluate  the  performance  of \textit{ScoreGAN}:
\begin{itemize}
    \item Maximum Number of Reviews (MNR): with a higher number of reviews written by a user in a  single day, the probability of the review to be fraud increases~\cite{Mukherjee2013}.
    \item Review Length (RL): reviews with a lower number of words are more probable to be fraud~\cite{Li2011}.
    \item Score Extremity (SE): reviews with low score = 1,2,3 are more probable to be fraud~\cite{Mukherjee2013}.
    \item Single Reviews (SR): if a review is the user’s sole review, it is more probable to be fraud~\cite{Shebuit2015}.
 
\end{itemize}
For the last two features, the value of the feature is 1 for probable fraud reviews and 0 otherwise. Note that the evaluation is performed on $D_g$ for this set of experiments. Fig \ref{fig:beh-yelp} and \ref{fig:beh-trip} display the performance of \textit{ScoreGAN} on different behavioral features. 

As Fig. \ref{fig:beh-yelp} and \ref{fig:beh-trip} show, \textit{WE + score} yields a better performance as compared with other feature combinations. Adding other features even results in a degradation in the performance of \textit{ScoreGAN}. Hence, using more features does not necessarily improve the performance of ScoreGAN. SR does not improve the performance, since most of the reviews are single and are the only review of a user. MNR is similar to SR in definition and shows a similar performance as SR. On the other hand, SE improves the performance and experimentally demonstrates what was stated in the Introduction regarding the importance of the score.  RL  also results in a lower performance (in terms of AP, AUC, and accuracy) as fraudsters manage to write long reviews to appear identical to readers. As a result, score-based features still perform as the top behavioral features in  fraud  detection,  because  the  primary  goal  of  a  fraudster is  to  promote/demote  an  item by managing  the  score  and sentiment of the review.  


\subsubsection{Ablative Study}
\label{sec:ablative}
To show the effectiveness of the new components employed in this study, we conducted various experiments on the ScoreGAN framework. The experiments include two sections; first, the effect of the score employed for both the generator and discriminator is examined, and then the effect of regularization is studied.\\
\textbf{\textit{Effect of Score:}}
In this section, we aimed to examine the importance of using the score in both the generator and discriminator. To prove the effectiveness of using the score in our proposed approach, we removed the score from both the discriminator and the generator, once for each and simultaneously to observe their effects on the performance. Note that to remove the score from the generator, we remove the regularization term ($\lambda L(G_\theta,Q)$ from Eq. \ref{eq:obj-fun-new}), so the generator generates the reviews without any constraint. However, to remove the score from the discriminator, we remove the score from the input word embeddings (Eq. \ref{eq:ReLU}) of the discriminator. 

\begin{figure*}
  \centering
  \includegraphics[width=\linewidth]{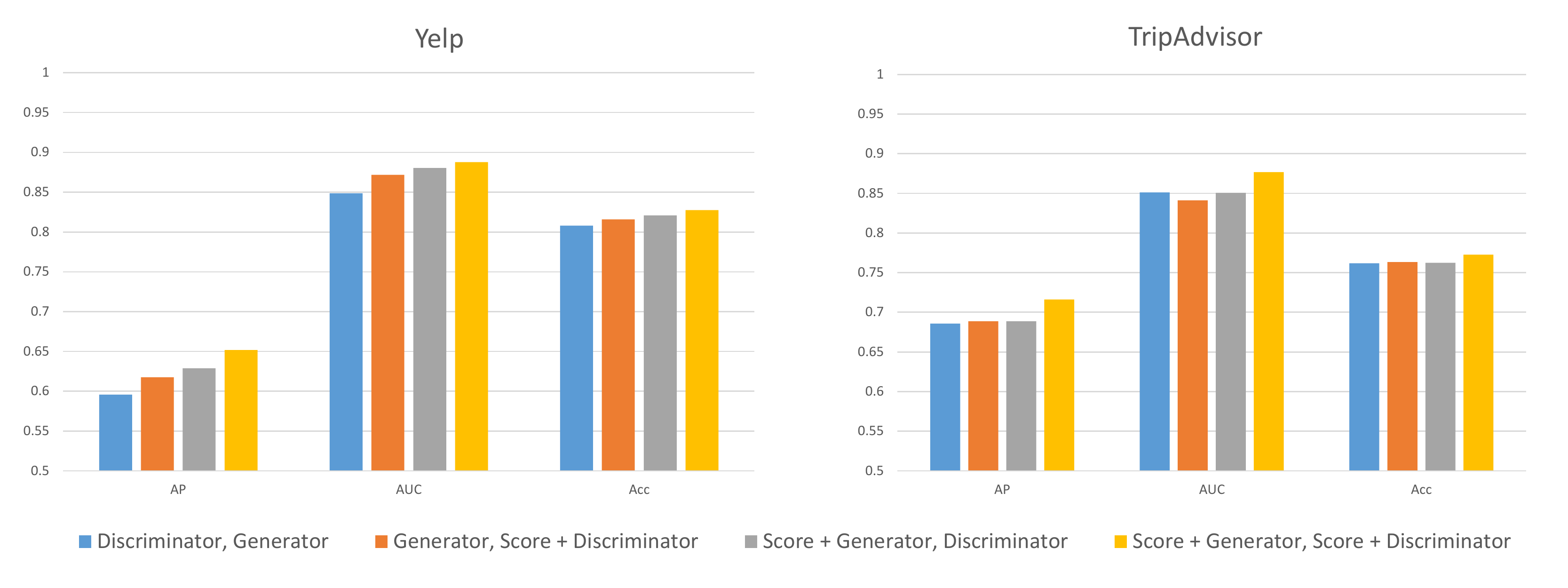}
  \caption{Effect of including the score in the $G$ and $D_g$ with 70\% of data as training set.}
\label{fig:ablative}
\end{figure*}

Fig. \ref{fig:ablative} represents the effect of using score in both the generator and discriminator on the performance of the proposed approach. Results on the Yelp dataset show that the performance is gradually improved after considering the score in the generator and the discriminator. Specifically, generating the reviews with the score has a greater impact on the performance as compared with including the score in the discriminator. To explain the improvement, one may say that generating the reviews correlated with score increases diversity in the generated data. More diverse data helps the discriminator to learn the model, more accurately. Including the score of the discriminator will also improve the performance for all three metrics, but clearly with less improvement in comparison with when the score is used in the generator. One simple conclusion is that increasing the training data in this task is effective in achieving a consistent final performance increasing pattern.
Conversely, the results on the Tripadvisor do not show such an increasing pattern. For AP, improvement is obvious given the score in both the generator and the discriminator. For Accuracy, the improvement is evident, however including the score in the generator, shows a negative impact. Degradation in performance could be the result of employing a binary score (like or dislike) in the Tripadvisor. Such a binary score prevents the accurate generation of reviews correlated with the score.\\ 
\textbf{\textit{Effect of Regularization:}}
One of the important issues of the standard GAN is instability. Different techniques exist to overcome this issue. Here we use regularization, as explained in Sec. \ref{secSub:gan-constraint} to improve the stability of the \textit{ScoreGAN}. A comparison study with two GAN-based approaches is conducted. Convergence is referred to a situation where the performance of an approach is stabilized after a specific number of iterations.

To examine convergence, 
we remove the regularization term ($\lambda L(G_\theta,Q)$ from Eq. \ref{eq:obj-fun-new}) from the objective function (Unregulated ScoreGAN) of $D_g$ to show the impact of regularization on the convergence.
We compare $AP$, $AUC$ of different systems with ScoreGAN in the sequential iterations of the adversarial learning. For pre-training, after 100 iterations, the convergence is achieved by the generator. This number is 50 for the discriminator. For the adversarial training step, it took 100 iterations to converge. 
\begin{figure*}
  \centering
  \includegraphics[width=\linewidth]{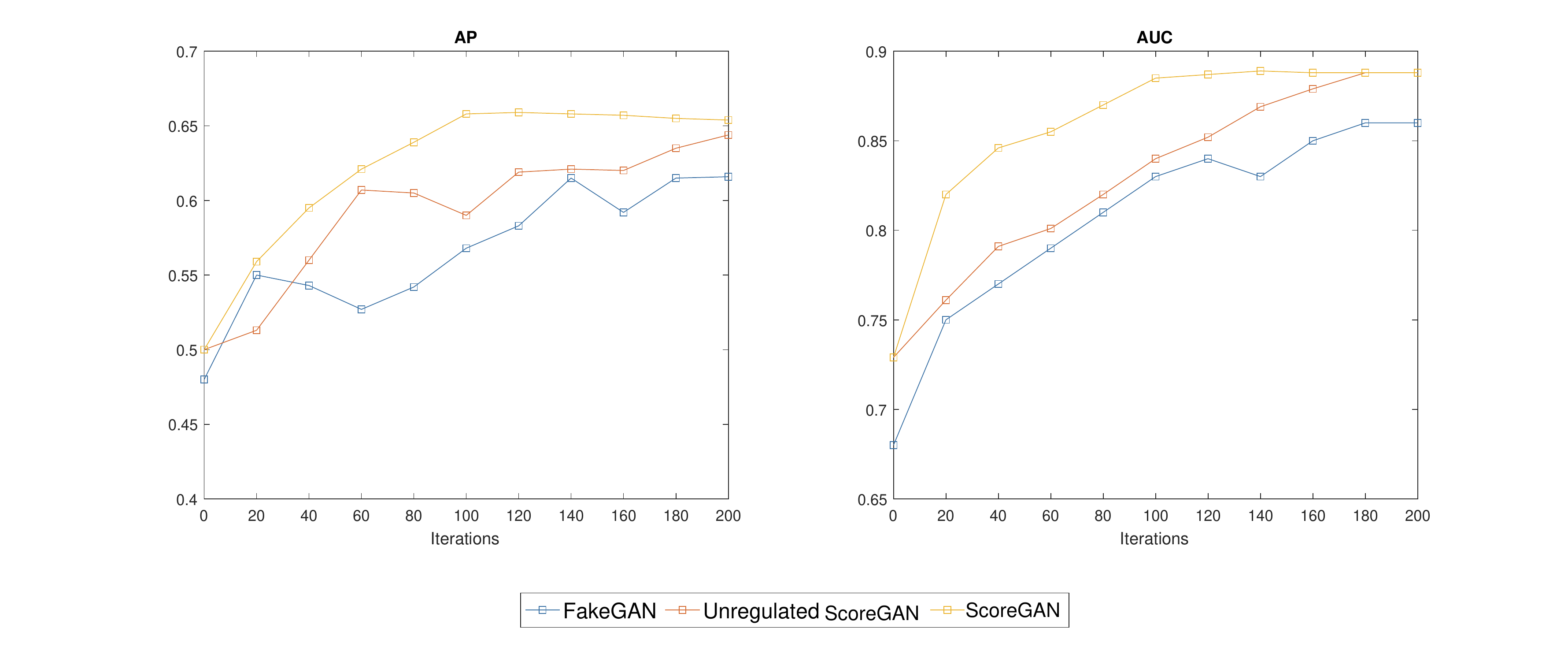}
  \caption{AP, AUC, and Accuracy for different supervisions of different frameworks on the Yelp dataset.}
    \label{fig:regularization}
\end{figure*}


Fig. \ref{fig:regularization} shows that for FakeGAN, 140 iterations are required to achieve the final performance, while for the unregulated ScoreGAN, the best performance is not achieved even after 200 iterations. Given the regularization term, ScoreGAN obtains convergence to the final performance in considerably faster 100 iterations.   

\subsubsection{Robustness with Data Scarcity}
\label{sec:robustness}
\begin{figure*}
  \centering
  \includegraphics[width=\linewidth]{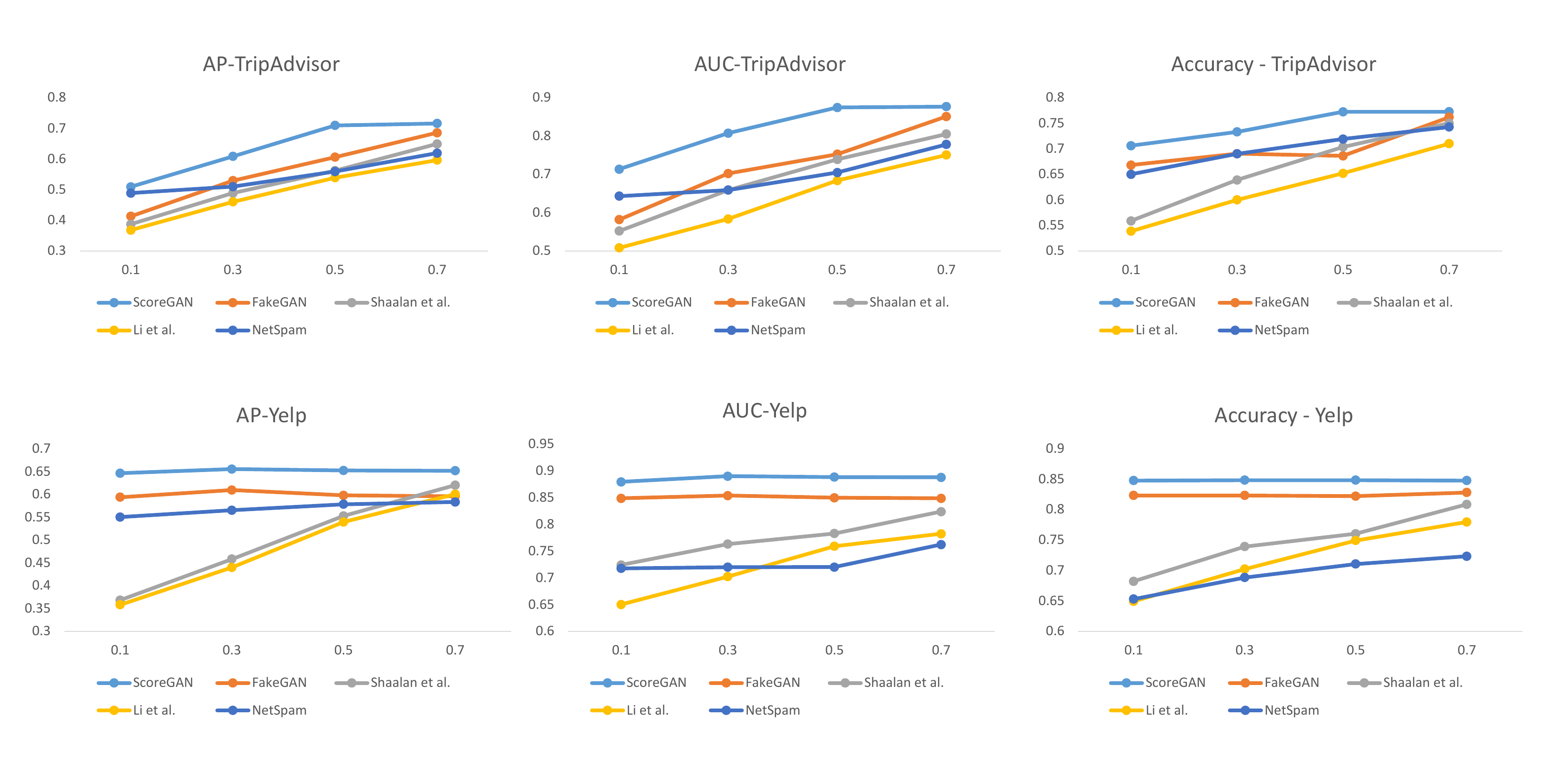}
  \caption{AP, AUC, and Accuracy for different supervisions for different frameworks.}
    \label{fig:robustness}
\end{figure*}

Due to data scarcity, robustness is considered to be an important matter in fraud review detection.
In this section, we conducted two sets of experiments, first to compare the performance of semi-supervised approaches against supervised approaches. In the next experiment, a different number of reviews of both datasets are selected and the cross-dataset performance of \textit{ScoreGAN} is compared with the two other semi-supervised frameworks.\\
\textit{\textbf{Robustness to Supervision Levels:}}
We partitioned the main dataset into a train and test dataset with different ratios and refer to these different partitions as ``supervisions" (0.7, 0.5, 0.3, and 0.1 as training set, respectively, and the remaining as test set). Hence, we use different supervision levels (proportion of data) to demonstrate the robustness of \textit{ScoreGAN}. Fig. \ref{fig:robustness} 
shows the results against supervision levels on the \textit{ScoreGAN} compared with two semi-supervised approaches (\textit{NetSpam}, \textit{FakeGAN}) and two supervised approaches (Li \textit{et al.}~\cite{li2020detection}, Shaalan \textit{et al.}~\cite{shaalandetecting}).
Fig. \ref{fig:robustness} shows that \textit{ScoreGAN} is robust to data scarcity for all three metrics, and this can be observed by the effects of different supervision levels. 
For $AP$, the best result is obtained with 0.7 supervision for TripAdvisor, while the performance for 0.5 of data as the training set remains the same. \textit{ScoreGAN} performs similarly on the Yelp dataset and converges to a constant value with 30\% supervision. 
In addition, the variations in $AP$ is more than the variations of $AUC$ or $Accuracy$, especially for \textit{ScoreGAN}. This happens because the values for $AP$ are relatively low and a slight change in the amount of training data, leads to a larger change in performance as compared with other measures. For $AUC$, the rate of improvement in \textit{ScoreGAN} decreases against the supervisions. \textit{NetSpam} works better than \textit{FakeGAN} with $10\%$ data on TripAdvisor, but as the amount of samples increases, \textit{FakeGAN} becomes superior, while \textit{ScoreGAN} performs considerably the best for all of the supervision levels. In addition, the performance of \textit{ScoreGAN} ensures a convergence point for all three metrics on Yelp dataset (five times larger than TripAdvisor).
This shows how important the amount of data is for deep learning approaches. 
The results also indicate that with lower levels of supervision (less training data), the performance of two supervised approaches (Li \textit{et al.}~\cite{li2020detection} and Shaalan \textit{et al.}~\cite{shaalandetecting}) decrease, drastically. On the other hand,  the  performance  of  ScoreGAN and the other two semi-supervised approaches is less affected by the proportion of the training data.
Finally, \textit{ScoreGAN} demonstrates promising scalable results in terms of accuracy. We attribute this to \textit{ScoreGAN}'s ability in generating bot reviews for data augmentation purposes. \\ 
\textit{\textbf{Cross-dataset Robustness:}}
To investigate the performance of ScoreGAN compared to the two other semi-supervised approaches on small datasets, we also conducted a cross-dataset evaluation of robustness. Cross-dataset evaluations can guarantee the performance of our proposed approach on different domains, while such evaluation also shows that data augmentation will result in gaining fair performance even with a small training dataset. In this experiment, a different number of samples are selected from each dataset, and the performance of the frameworks are compared against each other based on the same number of samples. We call this process the ``cross-dataset" experiment. 
\begin{figure*}
  \centering
  \includegraphics[width=\linewidth]{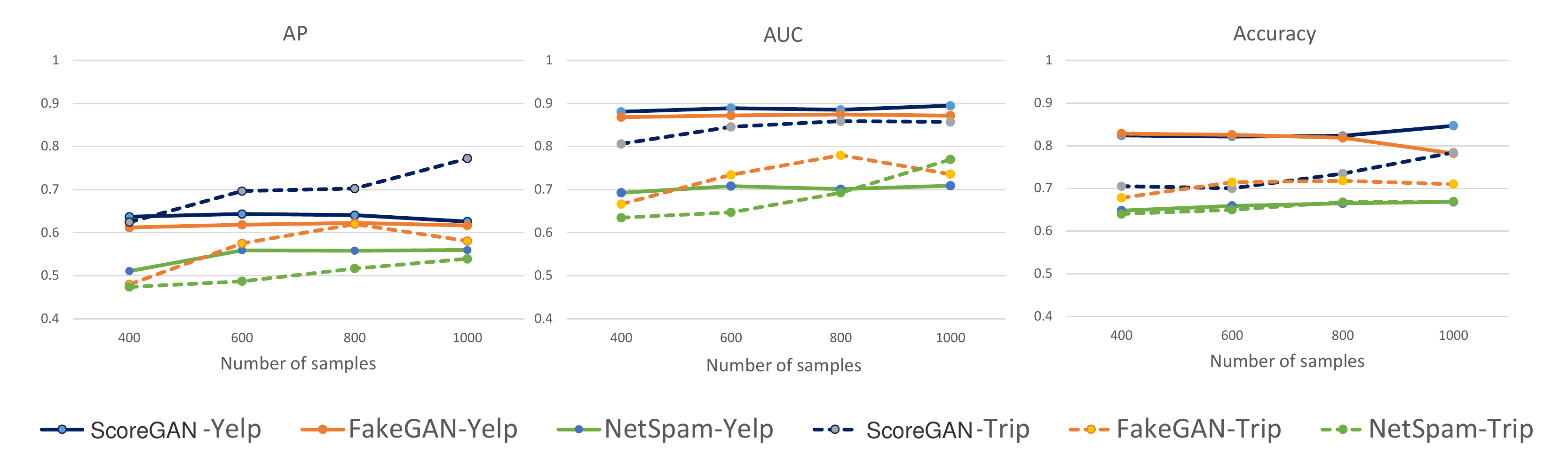}
  \caption{AP, AUC, and Accuracy for cross-dataset experiment on different frameworks.}
    \label{fig:robustness-review-no}
\end{figure*}
Fig. \ref{fig:robustness-review-no} represents the cross-dataset performance of all frameworks on both datasets. Generally, the performance of ScoreGAN is consistent with the number of samples for the Yelp dataset. On the other hand, the performance of ScoreGAN improves on TripAdvisor, given more samples from the training set. The fluctuation for other frameworks reflects their sensitivity to the training data proportion. Anyway, the robustness of the frameworks decays with the small number of reviews in the training dataset. However, the performance of ScoreGAN exhibits imperceptible changes when compared with other frameworks.

Fig.~\ref{fig:robustness-review-no} shows that with more samples from TripAdvisor the performance of FakeGAN is degraded in terms of AP, AUC, and accuracy, while ScoreGAN is improved for the three metrics. The main reason is that TripAdvisor provides samples that are evenly balanced on fraud/genuine reviews and with more data the generator is provided with more fraud reviews to generate more diverse fraud reviews. 
For the Yelp dataset, the performance of ScoreGAN does not fluctuate except for AP, which is slightly decreased with 1000 samples. ScoreGAN is improved with 1000 samples in terms of AUC and accuracy.  This is also because the Yelp dataset contains significantly more genuine reviews than fraud reviews. Hence, the generator is provided with fewer fraud reviews in the training step and the performance of the discriminator remains constant with little fluctuations. 

\section{Conclusion}
In this paper, we proposed \textit{ScoreGAN}, a regulated GAN with one generator and dual discriminators for fraud review detection that is capable of making use of both the review text and metadata, such as scores. Information gain maximization between the score and the generated score-correlated review is used as the basic idea for a new loss function, which will not only stabilize the GAN, addressing the low convergence issue, but also focus the Generator to automatically produce more human like bot reviews. 
The \textit{ScoreGAN} produced AUC of 88.78\% and AP of 65.16\% on the Yelp dataset which is a significant improvement over what \textit{FakeGAN} \cite{Aghakhani2018}, \textit{NetSpam} \cite{Shehnepoor2017}, Li \textit{et al.}~\cite{li2020detection}, and Shaalan \textit{et al.}~\cite{shaalandetecting}. Future work will focus on fraud review detection considering a combination of text features with other features which affect the performance. This can also be helpful for acquiring a joint representation of both text and metadata. 

\bibliographystyle{IEEEtran}
\bibliography{References}

\end{document}